\documentclass[twoside,11pt]{article}

\usepackage{footmisc}
\usepackage{jmlr2e}
\usepackage[utf8]{inputenc}
\usepackage{tabularx}
\usepackage{xcolor}

\usepackage[scaled]{beramono}
\usepackage{listings}
\usepackage{wrapfig}
\usepackage{capt-of}

\usepackage{todonotes}

\DeclareFixedFont{\ttb}{T1}{txtt}{bx}{n}{9}
\DeclareFixedFont{\ttm}{T1}{txtt}{m}{n}{9}

\usepackage{color}
\definecolor{deepblue}{rgb}{0,0,0.5}
\definecolor{deepred}{rgb}{0.6,0,0}
\definecolor{deepgreen}{rgb}{0,0.5,0}

\hypersetup{pdftitle={OpenML-Python: an extensible Python API for OpenML},pdfauthor={Matthias Feurer, Jan N. van Rijn, Arlind Kadra, Pieter Gijsbers, Neeratyoy Mallik, Sahithya Ravi, Andreas Müller, Joaquin Vanschoren, Frank Hutter}}

\usepackage{lastpage}
\jmlrheading{22}{2021}{1-\pageref{LastPage}}{11/19}{5/21}{19-920}{Matthias Feurer, Jan N. van Rijn, Arlind Kadra, Pieter Gijsbers, Neeratyoy Mallik, Sahithya Ravi, Andreas Müller, Joaquin Vanschoren and Frank Hutter}

\ShortHeadings{OpenML-Python: an extensible Python API for OpenML}{Feurer et al.}
\firstpageno{1}

\begin{document}

\title{OpenML-Python: an extensible Python API for OpenML}

\author{\name Matthias Feurer \email feurerm@cs.uni-freiburg.de \\
    \addr University of Freiburg, Freiburg, Germany
    \AND
    \name Jan N. van Rijn \email j.n.van.rijn@liacs.leidenuniv.nl \\
    \addr Leiden University, Leiden, Netherlands
    \AND
    \name Arlind Kadra \email kadraa@cs.uni-freiburg.de\\
    \addr University of Freiburg, Freiburg, Germany
    \AND
    \name Pieter Gijsbers \email p.gijsbers@tue.nl \\
    \addr Eindhoven University of Technology, Eindhoven, Netherlands
    \AND
    \name Neeratyoy Mallik \email mallik@cs.uni-freiburg.de \\
    \addr University of Freiburg, Freiburg, Germany
    \AND
    \name Sahithya Ravi \email s.ravi@tue.nl \\
    \addr Eindhoven University of Technology, Eindhoven, Netherlands
    \AND
    \name Andreas Müller \email andreas.mueller.ml@gmail.com \\
    \addr Microsoft, Sunnyvale, USA
    \AND
    \name Joaquin Vanschoren \email j.vanschoren@tue.nl \\
    \addr Eindhoven University of Technology, Eindhoven, Netherlands
    \AND
    \name Frank Hutter \email fh@cs.uni-freiburg.de \\
    \addr University of Freiburg \& Bosch Center for Artificial Intelligence, Freiburg, Germany
}

\editor{Balazs Kegl}

\maketitle

\begin{abstract}%
OpenML is an online platform for open science collaboration in machine learning, used to share datasets and results of machine learning experiments.
In this paper, we introduce \emph{OpenML-Python}, a client API for Python, which opens up the OpenML platform for a wide range of Python-based machine learning tools.
It provides easy access to all datasets, tasks and experiments on OpenML from within Python.
It also provides functionality to conduct machine learning experiments, upload the results to OpenML, and reproduce results which are stored on OpenML.
Furthermore, it comes with a scikit-learn extension and an extension mechanism to easily integrate other machine learning libraries written in Python into the OpenML ecosystem. Source code and documentation are available at \url{https://github.com/openml/openml-python/}.
\end{abstract}


\section{Introduction}
OpenML is a collaborative online machine learning (ML) platform, meant for sharing and building on prior empirical machine learning research~\citep{vanschoren-sigkdd14a-local}. 

It goes beyond open data repositories, such as 
UCI~\citep{UCI_2019}, PMLB~\citep{olson-bdm2017a_short}, the `datasets' submodules in scikit-learn and tensorflow \citep{scikit-learn_short,tensorflow_short}, and the closed-source data sharing platform at Kaggle.com, since OpenML also collects millions of shared experiments on these datasets, linked to the exact ML pipelines and hyperparameter settings, and includes comprehensive logging and uploading functionalities which can be accessed programmatically via a REST API. 
An introduction and detailed information can be found on \url{https://docs.openml.org}.

OpenML-Python is a seamless integration of OpenML into the popular Python ML ecosystem,\footnote{\url{https://github.blog/2019-01-24-the-state-of-the-octoverse-machine-learning/}} that takes away this complexity by providing easy programmatic access to all OpenML data and by automating the sharing of new experiments.\footnote{Other clients already exist for R~\citep{casalicchio-2017a} and Java~\citep{Rijn2016}.}
In this paper, we introduce OpenML-Python's core design, showcase its extensibility to new ML libraries, and give code examples for several common research tasks.

\section{Use cases for the OpenML-Python API}
OpenML-Python allows for easy dataset and experiment sharing and reuse by
handling all communication with OpenML's REST API.
In this section, we briefly describe how the package can be used in several common machine learning tasks and highlight recent uses.

\vspace*{0.2cm}\noindent{}\textbf{Working with datasets.}
OpenML-Python can retrieve the thousands of datasets on OpenML (all of them, or specific subsets) in a unified format, retrieve meta-data describing them, and search through them with filters.
Datasets are converted from OpenML's internal format
into \emph{numpy}, \emph{scipy} or \emph{pandas} data structures, which are standard for ML in Python.
To facilitate contributions from the community, it allows people to upload new datasets in only two function calls, and to define new \textit{tasks} on them (combinations of a dataset, train/test split and target attribute). 

\vspace*{0.2cm}\noindent{}\textbf{Publishing and retrieving results.}
Sharing empirical results allows anyone to search and download them in order to reproduce and reuse them in their own research.
One goal of OpenML is to simplify the comparison of new algorithms and implementations to existing approaches by comparing to the results on OpenML.
To this end we also provide an interface for integrating new machine learning libraries with OpenML and we have already integrated \emph{scikit-learn}.
OpenML-Python can then be used to set up and conduct machine learning experiments for a given \emph{task} and \emph{flow} (an ML pipeline), and publish reproducible results (including hyperparameter settings and random states).

\vspace*{0.2cm}\noindent{}\textbf{Use cases in published works.} OpenML-Python has already been used to scale up studies with hundreds of consistently formatted datasets~\citep{feurer-nips2015a,fusi-neurips2018a}, supply large amounts of meta-data for meta-learning~\citep{perrone-nips18a}, answer questions about algorithms such as hyperparameter importance~\citep{rijn-kdd18a} and facilitate large-scale comparisons of algorithms~\citep{strang_ada18a}.

\section{High-level Design of OpenML-Python}
The OpenML platform is organized around several entity types which describe different aspects of a machine learning study.
It hosts \emph{datasets}, \emph{tasks} that define how models should be evaluated on them, \emph{flows} that record the structure and other details of ML pipelines, and \emph{runs} that record the experiments evaluating specific \emph{flows} on certain \emph{tasks}.
For instance, an experiment (\emph{run}) shared on OpenML can show how a random forest (\emph{flow}) performs on `Iris' (\emph{dataset}) if evaluated with 10-fold cross-validation (\emph{task}), and how to reproduce that result.
In OpenML-Python, all these entities are represented by classes, each defined in their own submodule.
This implements a natural mapping from OpenML concepts to Python objects.
While OpenML is an online platform, we facilitate offline usage as well.

\noindent{}\textbf{Extensions.}
To allow users to automatically run and share machine learning experiments with different libraries 
through the same OpenML-Python interface, we designed an extension interface that standardizes the interaction between machine learning library code and OpenML-Python.
We also created an extension for \emph{scikit-learn}~\citep{scikit-learn_short}, as it is one of the most popular Python machine learning libraries.
This extension can be used for any library which follows the \emph{scikit-learn} API~\citep{sklearn_api_short}.

An extension's responsibility is to convert between the libraries' models and OpenML flows, interact with its training interface and format predictions. For example, the \emph{scikit-learn} extension can convert an \emph{OpenMLFlow} to an Estimator (including hyperparameter settings), train models and produce predictions for a task, and create an \emph{OpenMLRun} object to upload the predictions to the OpenML server.
The 
extension also handles advanced procedures, such as  \emph{scikit-learn}'s random search or grid search and uploading its traces (hyperparameters and scores of each model evaluated\\ during search).
We are working on more extensions, and\\ anyone can contribute their own using the \emph{scikit-learn}\\ extension implementation as a reference.\\\vspace*{-1.4cm}
\begin{wrapfigure}[9]{r}{0.4\textwidth}
\vskip -0.52in
  \centering
  \includegraphics[width=0.35\textwidth]{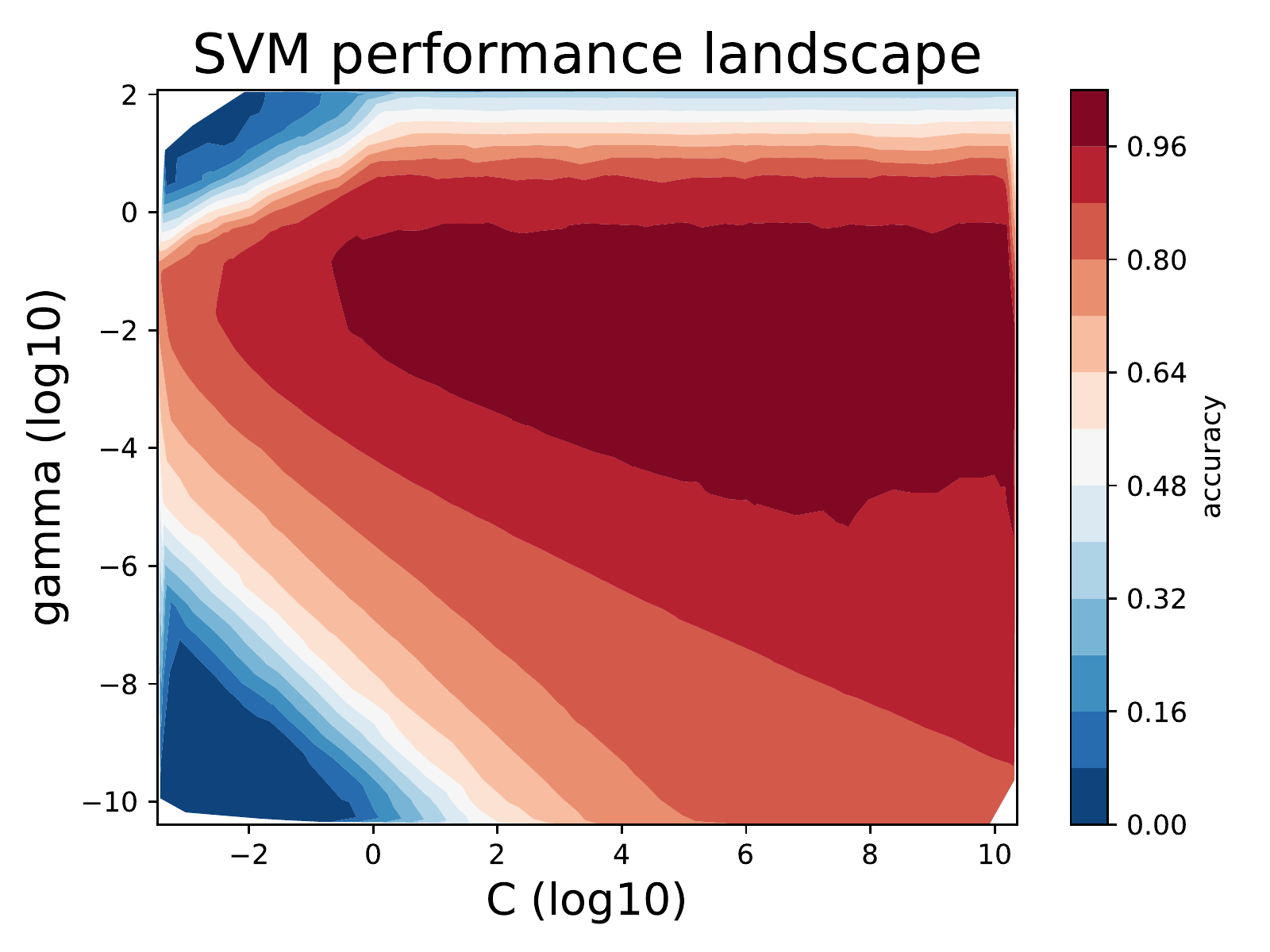}\\
  \footnotesize{SVM hyperparameter contour plot generated 
  by the code in Figure \ref{alg:plot_hyperparameter_contours}.}
  \label{fig:bohb_pcdarts_gdas}
\end{wrapfigure}

\begin{figure}[t]
\begin{footnotesize}
\begin{lstlisting}[language=Python]
import openml; import numpy as np
import matplotlib.pyplot as plt
df = openml.evaluations.list_evaluations_setups(
    'predictive_accuracy', flows=[8353], tasks=[6],
    output_format='dataframe', parameters_in_separate_columns=True,
) # Choose an SVM flow (e.g. 8353), and the dataset 'letter' (task 6).
hp_names = ['sklearn.svm.classes.SVC(16)_C','sklearn.svm.classes.SVC(16)_gamma']
df[hp_names] = df[hp_names].astype(float).apply(np.log)
C, gamma, score = df[hp_names[0]], df[hp_names[1]], df['value']
cntr = plt.tricontourf(C, gamma, score, levels=12, cmap='RdBu_r')
plt.colorbar(cntr, label='accuracy')
plt.xlim((min(C), max(C))); plt.ylim((min(gamma), max(gamma)))
plt.xlabel('C (log10)', size=16); plt.ylabel('gamma (log10)', size=16)
plt.title('SVM performance landscape', size=20)
\end{lstlisting}
\vspace*{-0.1cm}
\caption{Code for retrieving the predictive accuracy of an SVM classifier on the `letter' dataset and creating a contour plot with the results.\label{alg:plot_hyperparameter_contours}}
\end{footnotesize}
\vspace*{-0.4cm}
\end{figure}

\section{Examples}
We show two example uses of OpenML-Python to demonstrate its API's simplicity. First, we show how to retrieve results and evaluations from the OpenML server in Figure \ref{alg:plot_hyperparameter_contours} (generating the plot on the right).
Second, in Figure \ref{alg:example_suite} we show how to conduct experiments on a benchmark suite~\citep{bischl-arxiv19a}.  
Further examples, including 
how to create datasets and tasks and how OpenML-Python was used in previous publications, can be found in the online documentation.\footnote{\label{fn1}We provide documentation, a list of extensions and code examples on  \url{http://openml.github.io/openml-python} and host the project on \url{http://github.com/openml/openml-python}.}

\begin{figure}[t]
\begin{footnotesize}
\begin{lstlisting}[language=Python]
from openml import study, tasks, runs, extensions
from sklearn import compose, impute, pipeline, preprocessing, tree
cont, cat = extensions.sklearn.cont, extensions.sklearn.cat # feature types 
clf = pipeline.make_pipeline( compose.make_column_transformer(
    (impute.SimpleImputer(), cont),
    (preprocessing.OneHotEncoder(handle_unknown='ignore'), cat)),
    tree.DecisionTreeClassifier())  # build a classification pipeline
benchmark_suite = study.get_suite('OpenML-CC18')  # task collection
for task_id in benchmark_suite.tasks:  # iterate over all tasks
  task = tasks.get_task(task_id)  # download the OpenML task
  run = runs.run_model_on_task(clf, task)  # run classifier on splits
  # run.publish()  # upload the run to the server; optional, requires API key
\end{lstlisting}
\vspace*{-0.2cm}
\caption{Training and evaluating a classification pipeline from scikit-learn on each task of the OpenML-CC18 benchmark suite~\citep{bischl-arxiv19a}.}
\label{alg:example_suite}
\end{footnotesize}
\vspace*{-0.4cm}
\end{figure}

\section{Project development}
The project has been set up for development through community effort from different research groups, and has received contributions from numerous individuals.
The package is developed publicly through Github which also provides an issue tracker for bug reports, feature requests and usage questions.
To ensure a coherent and robust code base we use continuous integration for Windows and Linux as well as automated type and style checking.
Documentation is also rendered on continuous integration servers and consists of a mix of tutorials, examples and API documentation.

For ease of use and stability, we use well-known and established 3rd-party packages where needed.
For instance, we build documentation using the popular \emph{sphinx Python documentation generator},\footnote{\url{http://www.sphinx-doc.org} \hspace{1cm} \textsuperscript{\label{footnote5}5}\url{https://sphinx-gallery.github.io/}} use an extension to automatically compile examples into documentation and Jupyter notebooks,\textsuperscript{\hyperref[footnote5]{5}} 
and employ standard open-source packages for scientific computing such as \emph{numpy}~\citep{numpy_short}, \emph{scipy}~\citep{scipy_short}, and \emph{pandas}~\citep{pandas}.
The package is written in Python3 and open-sourced with a 3-Clause BSD License.\footref{fn1}

\section{Conclusion}
\emph{OpenML-Python} allows easy interaction with OpenML from within Python.
It makes it easy for people to share and reuse the data, meta-data, and empirical results which are generated as part of an ML study. This allows for better reproducibility, simpler benchmarking and easier collaboration on ML projects.
Our software is shipped with a scikit-learn extension and has an extension mechanism to easily integrate other ML libraries written in Python.

\newpage

\small

\acks{\textbf{MF}, \textbf{NM} and \textbf{FH} acknowledge funding by the Robert Bosch GmbH. \textbf{AK}, \textbf{JvR} and \textbf{FH} acknowledge funding by the European Research Council (ERC) under the European Union’s Horizon 2020 research and innovation programme under grant no.\ 716721. \textbf{JV} and \textbf{PG} acknowledge funding by the Data Driven Discovery of Models (D3M) program run by DARPA and the Air Force Research Laboratory. \textbf{The authors} also thank Bilge Celik, Victor Gal and everyone listed at \url{https://github.com/openml/openml-python/graphs/contributors} for their contributions.}

\vskip 10pt

\setlength{\bibsep}{4pt}
\bibliography{shortstrings,lib,local,shortproc}

\end{document}